\documentclass[conference,compsoc]{IEEEtran}
\usepackage{multirow}
\usepackage[pdftex]{graphicx}
\ifCLASSOPTIONcompsoc
  \usepackage[nocompress]{cite}
\else
  \usepackage{cite}
\fi
\ifCLASSINFOpdf
\else
\fi
\begin{document}
%
\title{LEARNING A REPRESSION NETWORK FOR PRECISE VEHICLE SEARCH}

\author{\IEEEauthorblockN{
Qiantong Xu, Ke Yan, Yonghong Tian
}
\IEEEauthorblockA{
Institute of Digital Media\\
School of Electrical Engineering and Computer Science, Peking University\\
No.5 Yiheyuan Road, 100871, Beijing, China\\
\{xuqiantong, keyan, yhtian\}@pku.edu.cn
}
}


%


\maketitle

\begin{abstract}
The growing explosion in the use of surveillance cameras in public security highlights the importance of vehicle search from large-scale image databases. Precise vehicle search, aiming at finding out all instances for a given query vehicle image, is a challenging task as different vehicles will look very similar to each other if they share same visual attributes. To address this problem, we propose the \textit{Repression Network} (RepNet), a novel multi-task learning framework, to learn discriminative features for each vehicle image from both coarse-grained and fine-grained level simultaneously. Besides, benefited from the satisfactory accuracy of attribute classification, a bucket search method is proposed to reduce the retrieval time while still maintaining competitive performance. We conduct extensive experiments on the revised VehcileID \cite{liu2016deep} dataset. Experimental results show that our RepNet achieves the state-of-the-art performance and the bucket search method can reduce the retrieval time by about 24 times.
\end{abstract}


\ifCLASSOPTIONpeerreview
\begin{center} \bfseries EDICS Category: 3-BBND \end{center}
\fi
%
\IEEEpeerreviewmaketitle

\section{Introduction}
Nowadays, there is an explosive growing requirement of vehicle image retrieval and re-identification from large-scale surveillance image and video databases in public security systems. The license plate naturally is an unique ID of a vehicle, and license plate recognition has already been used widely in transportation management applications. Unfortunately, we can not identify a vehicle simply by its plate in some cases. First, some surveillance cameras are not designed for license plate capturing, i.e. the resolution of such cameras is not high enough to clearly show the numbers on license plate. Second, the performance of plate recognition systems decrease dramatically when they try to classify some confusing characters like ``8" and ``B", ``O" and ``0", ``D" and ``O", etc. Most importantly, license plates are often easily occluded, removed or even faked which makes license plates less relevant to each single vehicle. Therefore, visual attributes based precise vehicle retrieval, which aims at finding out the same vehicle across different surveillance camera views using information learned from vehicle appearance, has a great practical value in real-world applications. 

Though the problem of vehicle retrieval and re-identification has already been discussed for many years, most of the existing works rely on a various of different sensors \cite{datta2008image} and can only be used to retrieve vehicles sharing the same coarse-level attributes (e.g. color and model) rather than exactly the one in the query image. Compared with other popular retrieval problems like face and person retrieval, vehicle retrieval could be more challenging as many vehicles sharing one or more visual attributes have very similar visual appearance. In other words, even with classification results of coarse attributes, it is still not enough for us to know which exactly the vehicle is. However, there are some special marks such as customized paintings, decorations, or even scratches etc, that can be used to identify a vehicle from others. Therefore, precise vehicle retrieval algorithm should be able to not only capture coarse-grained attributes like color and model of each vehicle but also learn more discriminative feature representing unique details for it. 

\section{Related Work}
\begin{figure*}
  \center
  \includegraphics[width=14cm,height=7cm]{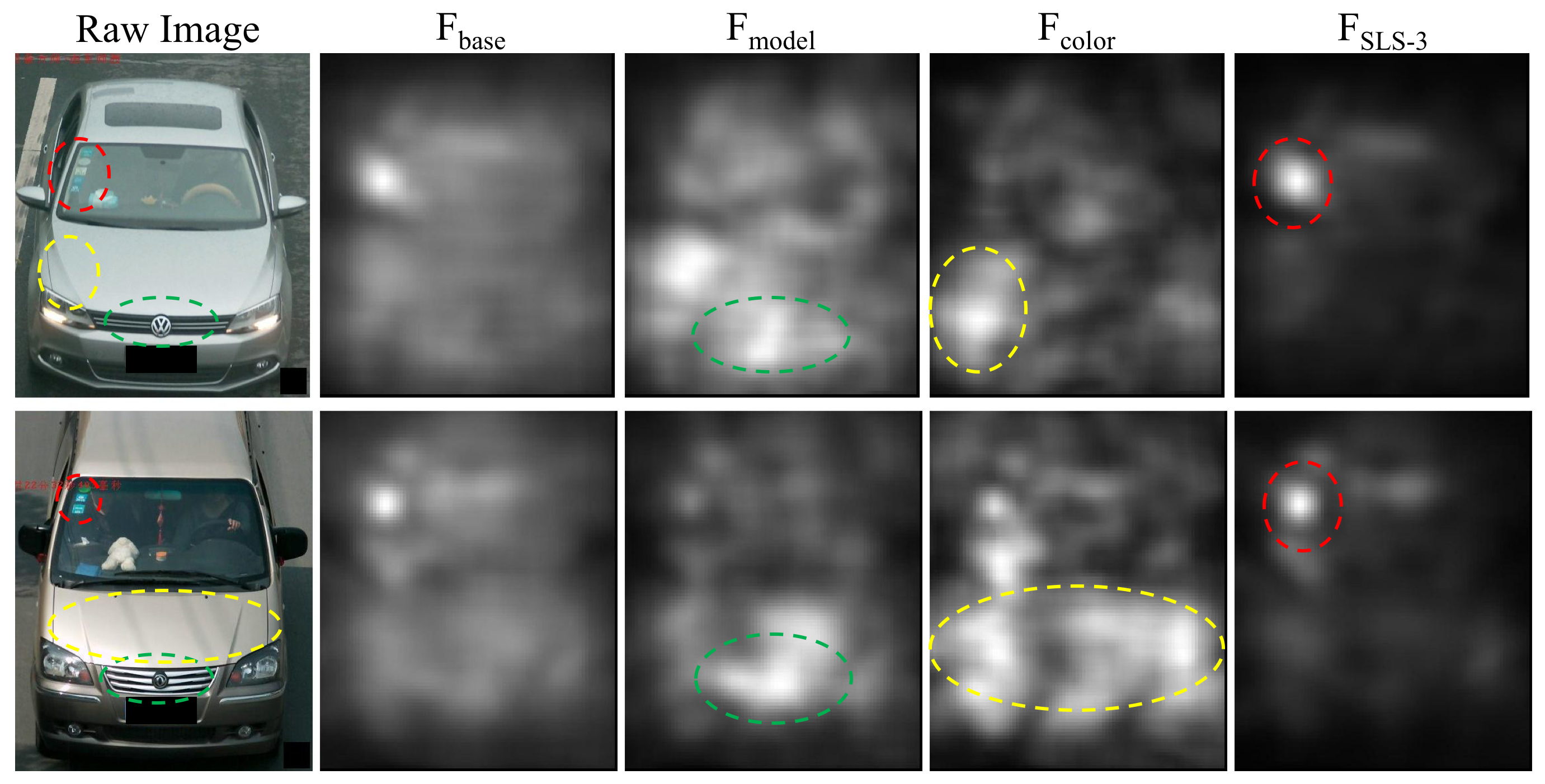}
  \caption{Saliency map of features from different layers in RepNet with PRL. The brighter an area is in the saliency map, the more information of the original image in that area is embedded in the given feature vector. The first column is the query image and column two to five shows the saliency maps of four feature vectors -- $F_{base}$, $F_{model}$, $F_{color}$ and $F_{SLS-3}$, respectively. }
   \label{fig:rf}
\end{figure*}

Recent advances in image understanding have been driven by the success of convolutional neural networks (CNN) \cite{krizhevsky2012imagenet} and many methods are proposed to improve the classification accuracy, by learning critical parts that can align the objects and discriminate between neighboring classes \cite{chai2013symbiotic, krizhevsky2012imagenet, zhang2013deformable}, or using distance metric learning to alleviate the issue of large intra-class variation \cite{qian2015fine, wang2014learning}. However, such models cannot perform well on fine-grained level tasks like precise retrieval which needs information beyond attribute labels. For example, a model can hardly tell the difference between vehicles having the same attributes (e.g. make, model, year and color) if it is only trained to classify those attributes. To get over this weakness, a potential solution is to incorporate similarity constraints (e.g. contrastive information \cite{chopra2005learning} or triplets \cite{chechik2010large}). This type of  constraints can effectively learn feature representations of details, which cannot be clearly described with attribute labels, for various tasks including verification and retrieval \cite{zhang2015embedding}. Note that triplet constraint has been widely used in feature learning for face recognition \cite{schroff2015facenet} and retrieval for many years, since it can preserve the intra-class variation. 

Therefore, an intuitive improvement is to combine the classification and the similarity constraints together to form a CNN based multi-task learning framework. For example, \cite{sun2014deep, yi2014learning} propose to combine the softmax and contrastive loss in CNN via joint optimization, while \cite{zhang2015embedding} combine softmax and triplet loss together to reach a better performance. These models improve traditional CNN because similarity constraints might augment the information for training the network. In addition, \cite{huang2015cross, zhao2015deep} concatenate features from different layers into a longer feature vector before being fed into loss layers so as to reduce information loss and increase the representation capability of intermediate layers. However, all these strategies still have several limitations. 1) They do not take the difference of each task into consideration, because all the visual features learned from the low-level convolutional groups and the following FC layers are shared among all the tasks. In this way, non-correlated constraints cannot be treated independently, harder tasks cannot be assigned with deeper layers (i.e. more weights) and interactions between constraints are implicit and uncontrollable.  2) Similarity constraints (e.g. triplet loss) as well as classification constraints (e.g. softmax loss) are all conducted directly on the last fully-connected feature, so that similarity constraints are partially constraining the same thing as classification constraints do. Since classification constraints are designed to learn features shared between vehicles in the same class, while similarity constraints is used to capture unique information of each vehicle, the undesired overlapping of the two kinds of constraints significantly limits the performance of similarity constraints by distracting them with those shared features. Besides, this overlapping can further hurt the convergence of model training. As shown in Figure \ref{fig:rf}, the saliency map of feature vectors generated by these models are very similar to those in the second column, which shows that the feature vectors embed information almost equally from every part of the vehicle. 

To address this problem, we propose a novel multi-task learning framework, RepNet, with two streams of fully-connected (FC) layers after convolutional groups for two different tasks -- attribute classification and similarity learning, and also a \textit{Repression Layer} to connect the two streams. So that the features generated from the relatively shallow stream, which is used for attribute classification, can be used as a feedback to guide the following similarity learning, to make them focus better on special details without embedding to much information of attributes. As shown in the third to fifth columns in Figure \ref{fig:rf}, different tasks do focus on different part of the vehicle and features learned from similarity constraints (col 5) have little correlation with those learned from classification constraints (col 3 and 4). 

In addition, benefited from the two stream structure of this framework, we introduce a bucket search strategy to significantly reduce the time we spend on retrieval without sacrificing too much accuracy. We also enrich the "VehicleID" dataset \cite{liu2016deep} with a new attribute and conduct extensive retrieval experimental on it. 

\section{The Proposed Approach}

\begin{figure*} \label{fig:model}
  \center
  \includegraphics[width=17cm,height=7cm]{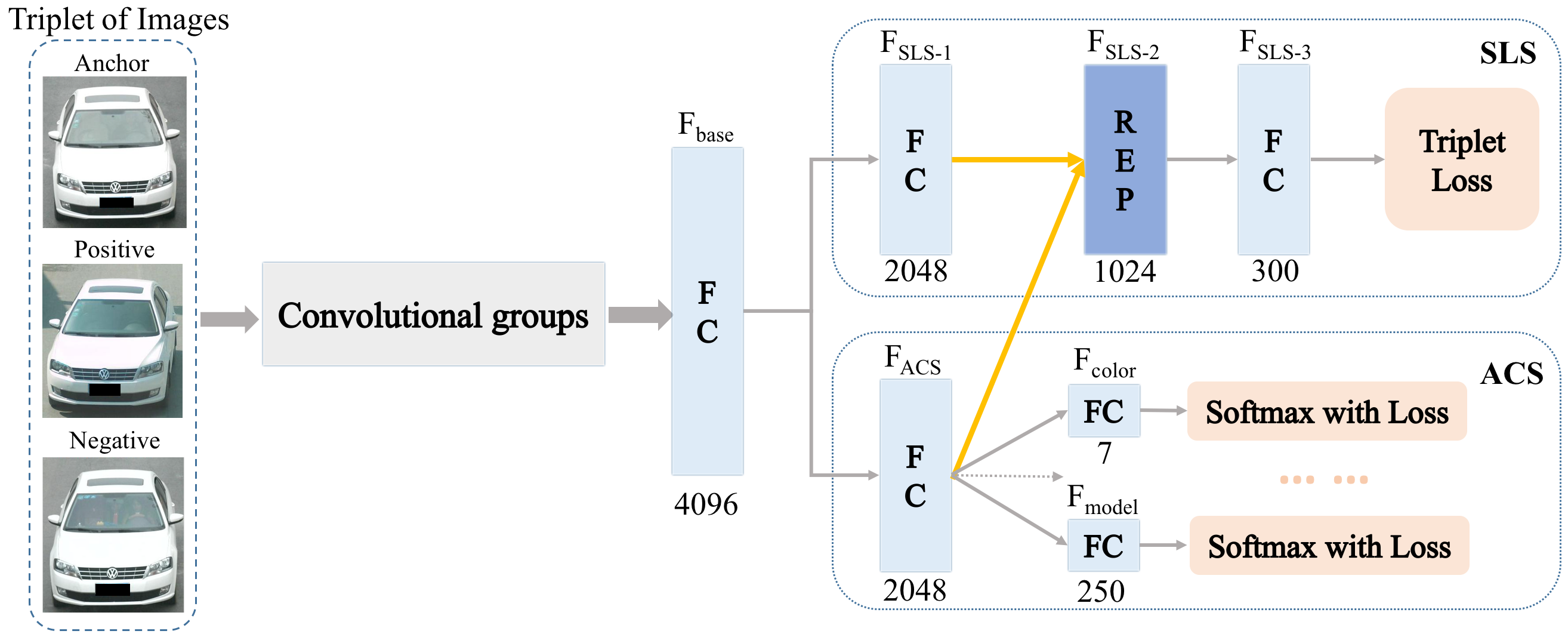}
  \caption{RepNet takes image triplets as input and feeds them through three networks, including convolutional groups, FC layers and Repression layers (REP), with shared weights. Names and dimensions of output features are listed above and under each layer. Repression layer takes two feature vectors $F_{SLS-1}$ and $F_{ACS}$ as input. Only the anchor image in each triplet is used for attribute classification, i.e. only the network for it has FC layers and loss functions after $F_{ACS}$.}
\end{figure*}

\subsection{Repression Network}
The main difference between other deep multi-task learning models and Repression Network is that we separate the fully-connected layers (FC layers) after the deep convolution structures into two streams, one for label-based attributes classification (Attributes Classification Stream, ACS) with softmax loss and one for detail-oriented similarity learning (Similarity Learning Stream, SLS) with triplet loss \cite{schroff2015facenet}. Then a Repression layer is added between the two streams. As shown in Figure \ref{fig:model}, the Repression layer connects the SLS with the ACS by interacting the feature from first layer in SLS ($F_{SLS-1}$) together with the feature from the final FC layer in ACS ($F_{ACS}$).

The basic idea of building such a model is that we want the deep network to 
generate two independent sub-features from two different levels -- coarse attributes and details, so that each sub-feature can embed more discriminative information for that level and can be better used to perform precise retrieval tasks. 
For example, if the query image is the top left one in Figure \ref{fig:rf}, we can first find all the images of grey Volkswagen Sagitar from the database as candidates and then narrow down using details like customized painting, decorations, scratches or other special marks. For vehicles, however, coarse-grained attributes and special marks are always independent, like a sticker or a scratch can appear on either a white KIA sedan or a black Honda SUV. That is why we can use two different streams for each of them. Moreover, based on previous experiment results\cite{zhang2015embedding}, coarse-grained attribute learning is much easier than the similarity learning, i.e. the former in a multi-task learning framework converges much faster than the latter and can reach a higher accuracy. Thus, we design the network with relatively fewer FC layers in ACS to get its final feature earlier. Finally, through the Repression layer, the early-extracted ACS feature is used to give certain feedback to the following SLS learning process, so as to reduce or even eliminate the information about those coarse-grained attributes being embedded into SLS and make it focus more on those potential details. 

In general, this is a novel multi-task learning framework which better utilizes the power of multi-task framework to generate two discriminative sub-features as independently as possible for two different level of tasks -- coarse-grained vehicle attributes classification and learning details beyond attribute labels. In addition, it is able to solve the issue of the slow and difficult convergence of the triplet loss in previous multi-task learning frameworks.

\subsection{Repression Layer}
The inputs of the Repression layer are two vectors -- $F_{ACS}$ and $F_{SLS-1}$, which have the same dimension denoted as $D_{input}$. The output of the Repression layer, denoted as $F_{SLS-2}$ with dimension $D_{ouput}$, is the feature vector fed into the second layer in SLS. We build three different Repression layers and compare them in the experiment. 

\textbf{Product Repression Layer (PRL)}. PRL performs element-wise product for two input feature vectors and maps the new vector into $D_{output}$ dimension space:
\begin{eqnarray}
F_{SLS-2} = W_{PRL}^{T}(F_{SLS-1} \circ F_{ACS}),
\end{eqnarray}
where $W_{PRL}$ is a $D_{input} \times D_{output}$ matrix. The derivatives of (1) with respect to each input feature vector are given by:
\begin{eqnarray}
\frac{\partial E}{\partial F_{SLS-1}} &=& (W_{PRL} \delta_{SLS-2}) \circ F_{ACS}, \\
\frac{\partial E}{\partial F_{ACS}} &=& (W_{PRL} \delta_{SLS-2}) \circ F_{SLS-1}, \\
\frac{\partial E}{\partial w_{ij}} &=&  \delta_{SLS-2}[j] \times F_{ACS}[i] \times F_{SLS-1}[i],
\end{eqnarray}
where $\delta_{SLS-2} = \frac{\partial E}{\partial F_{SLS-2}}$ is the gradient calculated for $F_{SLS-2}$, $w_{ij}$ is the element in the $i$th row and $j$th column of $W_{PRL}$ and $E$ represents the loss, which is triplet loss in our experiments, used for similarity learning.  

\textbf{Subtractive Repression Layer (SRL)}. Similarly, SRL performs element-wise subtraction for two input feature vectors and maps the new vector into $D_{output}$ dimension space:
\begin{eqnarray}
F_{SLS-2} = W_{SRL}^{T}(F_{SLS-1} - F_{ACS}),
\end{eqnarray}
where $W_{SRL}$ is also a $D_{input} \times D_{output}$ matrix. Likewise, the derivatives of (5) are given by:
\begin{eqnarray}
\frac{\partial E}{\partial F_{SLS-1}} &=& 
\frac{\partial E}{\partial F_{ACS}} ~=~~ W_{SRL} \delta_{SLS-2}, \\
\frac{\partial E}{\partial w_{ij}} &=&  \delta_{SLS-2}[j] \times (F_{ACS}[i] - F_{SLS-1}[i]).
\end{eqnarray} 

\textbf{Concatenated Repression Layer (CRL)}. CRL concatenates two input feature vectors into a larger vector and maps it into $D_{output}$ dimension space:
\begin{eqnarray}
F_{SLS-2} &=& W_{CRL}^{T}[F_{SLS-1}; F_{ACS}] \nonumber \\
~ &=& W_{CRL-1}^{T}F_{SLS-1} + W_{CRL-2}^{T}F_{ACS},
\end{eqnarray}
where $W_{CRL}$ is a $2D_{input} \times D_{output}$ matrix, $W_{CRL-1}$ and $W_{CRL-2}$ are $D_{input} \times D_{output}$ matrices. The derivatives of (8) are given by:
\begin{eqnarray}
\frac{\partial E}{\partial [F_{SLS-1}; F_{ACS}]} = W_{CRL} \delta_{SLS-2}, \\
\frac{\partial E}{\partial w_{ij}} =  \delta_{SLS-2}[j] \times [F_{SLS-1}; F_{ACS}][i].
\end{eqnarray}

Repression layer is designed to keep the information learned in ACS from being embedded into SLS and to balance the scale of weights in two streams. SRL is the most intuitive method to reduce information embedding from one feature to anther, which can be used as a baseline for RepNet. PRL increases the modeling capability as it is able to model the inner product of two feature vectors. We can assume two vectors have little correlation, if they are orthogonal to each other. Compared with the previous Repression layers which tend to directly model the relation of two features, CRL, a linear transformation and combination of two features, is the most magic one and is able to model more complicated cases beyond explicitly defined distance measurements.

\subsection{Training the Network}
We use cross-entropy loss on the softmax normalization for attributes classification and triplet loss for similarity learning. All the CNN models, based on VGG \cite{Simonyan14c} including five convolutional groups, in our experiments are trained with the widely-used deep learning framework "Caffe"\cite{jia2014caffe} and are fine-tuned from a base line model which can only be used for attribute classification. We set the mini-batch size for gradient descent to 90 which means $90 \times 3 = 270$ images are fed in each training iteration. We start with a base learning rate of 0.001 and then lower it by repeatedly multiplying 0.5 after every 50000 batch iterations. The momentum is set to 0.9 and weights for all losses are set equally.

In terms of similarity learning, one important issue is triplet sampling because it can directly determine what can be leaned by the network. For our Repression layer based multi-task learning framework, only the hardest triplets, in which the three images have exactly the same coarse-level attributes (e.g. color and model), can be used for similarity learning. The reason of doing so is that the similarity learning stream is designed to learn something beyond attributes, i.e. some details that can not be described by labels, so the features learned by SLS should have nothing to do with attributes classification. Otherwise, the feature learned from SLS could be partially overlapped with that from ACS, which will deteriorate the performance or expression of our model.

\section{Repression Analysis} \label{ssec:1}

\begin{table}[t]
\begin{center}
\caption{Correlation between $F_{SLS-1}$ and $F_{SLS-2}$} \label{tab:corr}
\begin{tabular}{|c|c|c|}
  \hline
  Model & Correlation & $p$-value\\
  \hline
  2S w/o REP & 0.9986 & $<10^{-2}$ \\
  RepNet+SRL & 0.9975 & $<10^{-2}$\\
  RepNet+CRL & 0.9953 & $<10^{-2}$\\
  RepNet+PRL & \textbf{0.9898} & $<10^{-2}$\\
  \hline
\end{tabular}
\end{center}
\end{table}

\subsection{Canonical Correlation Analysis}
We use Canonical Correlation Analysis (CCA) \cite{knapp1978canonical} to show that Repression layer can repress certain information being embedded in the following features, i.e. decrease the correlation between $F_{SLS-1}$ and $F_{SLS-2}$. Specifically, CCA will learn the projections for $F_{SLS-1}$ and $F_{SLS-2}$ features, through which the correlation between the projected features get maximized and this correlation is what we used to represent the relation between those two features. 

We first derive the covariance matrices $\Sigma_{XX}$ and $\Sigma_{YY}$ and cross-covariance matrices $\Sigma_{XY}$ and $\Sigma_{YX}$ for $F_{SLS-1}$ and $F_{SLS-2}$ which are denoted as $X$ and $Y$ on the training set. Then, without losing generality, we pick the eigen-vectors of the largest eigen-values
of $\Sigma _{XX}^{-1}\Sigma _{XY}\Sigma _{YY}^{-1}\Sigma _{YX}$ and $\Sigma _{YY}^{-1}\Sigma _{YX}\Sigma _{XX}^{-1}\Sigma _{XY}$ to project $F_{SLS-1}$ and $F_{SLS-2}$ onto a one dimensional space. The Pearson correlation coefficients between the projected features are listed in Table \ref{tab:corr}.
Note that the correlations are very close to 1 because the two features are only one layer apart and those values are the maximum correlation they can get by projecting to a one dimensional space. Nevertheless, we can still see a 1\% decrease of correlation (RepNet + PRL), which means there is certain information being repressed from passing into $F_{SLS-2}$ due to Repression layer. 

\subsection{Saliency Maps}
\begin{figure*}
  \center
  \includegraphics[width=17cm,height=9cm]{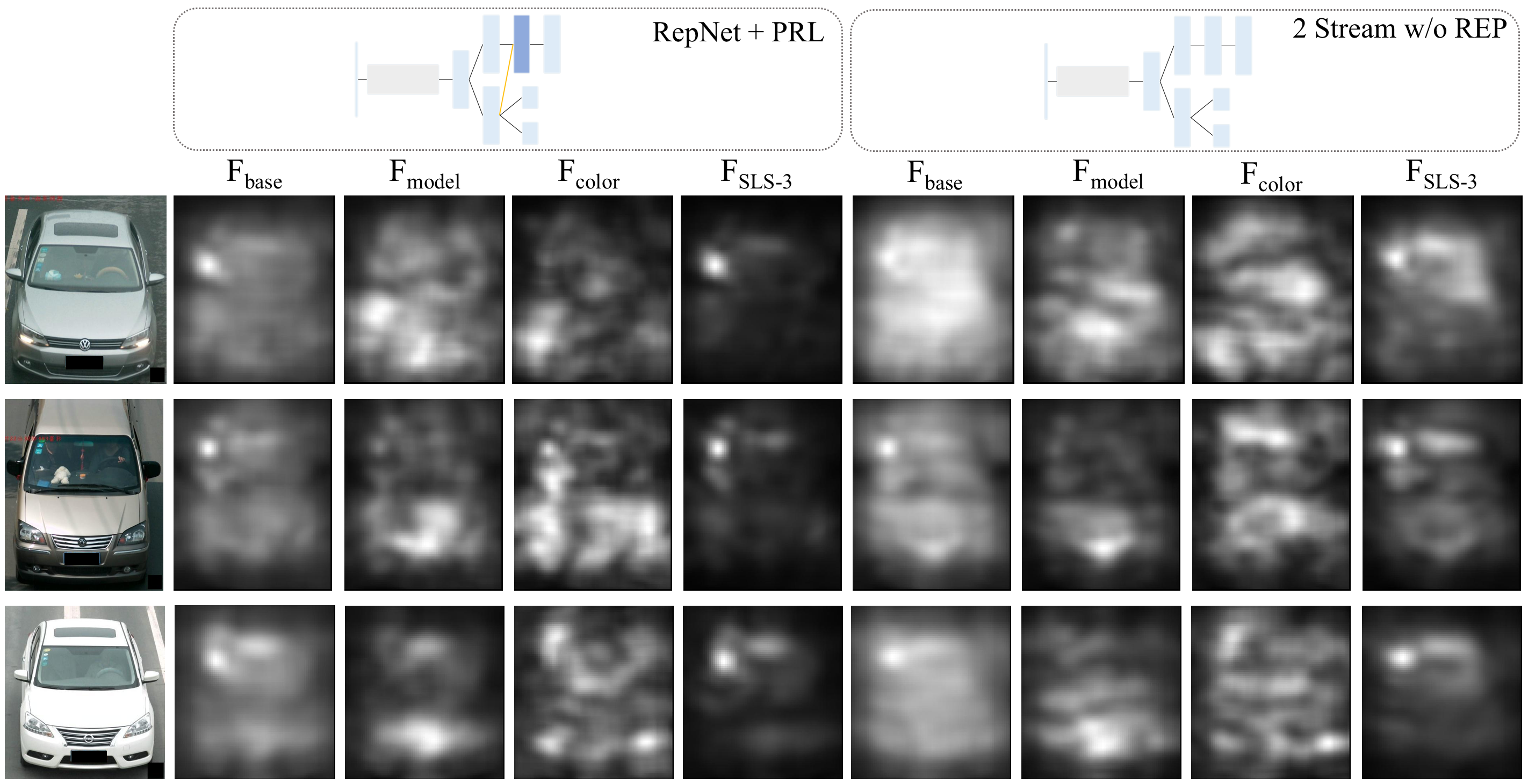}
  \caption{Saliency map of features from different layers in RepNet with PRL and a network with same architecture but repression layer. The first column is the query image and the following two groups of four columns shows the saliency maps of four feature vectors -- $F_{base}$, $F_{model}$, $F_{color}$ and $F_{SLS-3}$, respectively learned from two networks.}
   \label{fig:rf1}
\end{figure*}

We then demostrate the effectiveness of RepNet in a intuitive way. We use the method introduced in \cite{zeiler2014visualizing} to get the saliency maps of features extracted from different layers. 
Specifically, by covering the image with a small black square, we can calculate the difference between the features generated from the new image and from the original image. The larger the difference is, the more information RepNet gains from the covered area. The saliency map for the complete image can be generated by sliding the black square across the whole image. 

We take a sedan and a van as examples in Figure \ref{fig:rf} to show that different tasks have different saliency map, i.e. focus on different part of the vehicle. The saliency map of $F_{base}$ spreads almost uniformly over the vehicle in the image which means low-level layers extract information from almost everywhere of the vehicle. However, with strong constraints (i.e. loss functions) for each task, the saliency maps of features extracted from the final layer of each task fall on only some special area. For example, in order to correctly classify the model and color of the vehicle, our model focuses mainly on the front part of the vehicle, especially the brand on the cooling fin and the color on the engine hood. 

The saliency map of the last FC layer in SLS demonstrates the effectiveness of special detail extraction of our model and Repression layers. One reason is that it focuses on a small area on the top left of windshield on which all the inspection stickers are mandatory to put. Note that those stickers are compulsory for all the vehicles but can be put in a different order and pattern which vary from owner to owner. So, it can be used as a strong and special mark to distinguish the vehicle in the current query image from other vehicles having the same color and model. 
The other reason is that, as shown in Figure \ref{fig:rf1}, compared with the saliency map of the last FC layer in SLS of the network without Repression layer (col 9), the one for the RepNet (col 5) includes much less information from other parts of the vehicle, especially that from the front part which is used for attributes classification. This is strong evidence that Repression layer does repress the information used for other tasks and makes the current stream focus on learning the details beyond attribute labels.

\section{Retrieval Experiments}
In this section, we conduct thorough experiments of this proposed model on the ``VehicleID" dataset\cite{liu2016deep}, a fine-grained dataset of vehicle images with label structures. Particularly, we aim to demonstrate that our learned feature representations can be used to retrieve images of the same instance with significantly higher precision or less retrieval time than other CNN-based multi-task learning methods. Note that we do not introduce any single-task models (e.g. softmax classification or triplet learning) here because \cite{zhang2015embedding} has shown the effectiveness of multi-task framework over those models in discriminative features learning. 

The four state-of-art multi-task learning models we compare with are: 1) one-stream model with sotfmax and triplet loss on last FC layer \cite{zhang2015embedding}, 2) one-stream model with sotfmax and triplet loss on concatenated FC layers \cite{huang2015cross}, 3) two-stream model with softmax and triplet loss at the end of each stream, 4) two-stream model with softmax and coupled clusters loss, i.e. Mixed Difference Network \cite{liu2016deep}. All the CNN models have exactly the same five convolution groups at the bottom and are fine-tuned on the same training data.

\subsection{Dataset}
The VehicleID dataset contains data captured during daytime by multiple real-world surveillance cameras distributed in a small city in China. It contains 
221,763 images of 26,267 vehicles in total and 78,982 images, which have been used for model training, are labeled with three attributes -- color, vehicle model and vehicle ID. Color and vehicle model are used as two coarse-grained attributes in our experiments and there are 7 distinct colors, 250 distinct models in total. For each vehicle ID, i.e. each vehicle class, there are more than one images belonging to that class. Thus, VehicleID dataset is suitable for our model training and retrieval experiments. Additionally, in our experiments, we also include 389,316 distractors which are vehicle images collected by surveillance cameras in another city.

However, one significant weakness of VehicleID dataset is that the vehicle images are either captured from the front or the back and there is no label about this information. So, we give all the images in VehicleID dataset a new label \textit{view} to describe from where the images were taken. Then, we update the vehicle ID for each image according to its \textit{view}, so that images in the same original class but with different \textit{view}s are split into two new classes. Now, we have 43,426 new distinct vehicles IDs. 

We create two lists of query images from VehicleID dataset for our retrieval experiments. Note that all those query images do not appear in the training set and there is no pair of them sharing the same ID. The first query list, \textit{Random Query List}, is made up of randomly selected 1000 images while the other one, \textit{Tough Query List}, includes 1000 images with the largest inner-class sample number. 

\subsection{Evaluation Criteria}
We evaluate our proposed model on the newly relabeled VehicleID dataset following the widely used protocol in object retrieval -- mean average precision(MAP) and we also report the ``precision at k" curves. 

``Precision at k" denotes the average precision for the top $k$ retrieved images,
$precision@k = \frac{1}{k} \sum_{i=1}^{k} \Pi (r_{i}=r_{q})$,
where $\Pi(\cdot) \in \{0,1\}$ is an indicator function, $r_{i}$ and $r_{q}$ is the label of $i$th image in the ranking list and the label of the query image. MAP is a more comprehensive measurement describing precision and recall at the same time:
$MAP = \frac{1}{T} \sum_{k=1}^{N} \Pi (r_{k}=r_{q})precision@k$,
where $T$ is the ground truth number of the query image, $N$ is the total number of images in the database.

\subsection{Bucket Search}

Compared to the linear search method which exhaustively computes the distance between query image and all the images in the database, the bucket search only does this for a list of candidate neighbors. Specifically, we first create several buckets for each distinct attribute combination and put all the images into its corresponding bucket. Note that attributes of images in database, except those in the training set, and query image are predicted by the model. Then, for each query image, we find out the bucket it belongs to according to its predicted attributes and only compute the distance of features generated from SLS, which is a much shorter vector, between the query image and images lie inside the same bucket. But for one stream models, we still use the original feature for bucket search in our experiments.

In practice, however, since attributes classification cannot work perfectly, images may not be assigned to the correct bucket. We decide to check more than one buckets for each query image. Specifically, for VehicleID dataset, we check images in four buckets which represent the nearest two colors and two models of the given query image. 

\subsection{Experiment Results} 

\begin{table}[t]
\begin{center}
\caption{MAP of Vehicle Retrieval Task} \label{tab:map}
\begin{tabular}{|c|c|c|c|c|}
  \hline
  \multirow{3}{*}{Model} & \multicolumn{2}{|c|}{Random Query List} & \multicolumn{2}{|c|}{Tough Query List}\\
  \cline{2-5}
   & Linear & Bucket  & Linear & Bucket  \\
   & Search & Search & Search & Search
  \\
  \hline
  1~Stream\cite{zhang2015embedding} & 0.328 & 0.269 & 0.327 & 0.254 \\
  1S+Conc-FC\cite{huang2015cross} & 0.325 & 0.265 & 0.334 & 0.250 \\
  2S w/o REP & 0.291 & 0.315 & 0.299 & 0.268 \\
  2S+CCL w/o REP\cite{liu2016deep} & 0.315 & 0.279 & 0.297 & 0.255 \\
   \hline
  RepNet+SRL & 0.328 & 0.308 & 0.344 & 0.277\\
  RepNet+PRL & 0.402 & 0.305 & 0.432 & 0.283 \\
  RepNet+CRL & \textbf{0.441} & \textbf{0.341} & \textbf{0.457} & \textbf{0.288} \\
  \hline
\end{tabular}
\end{center}
\end{table}

\begin{figure}
  \center
  \includegraphics[width=8cm,height=7cm]{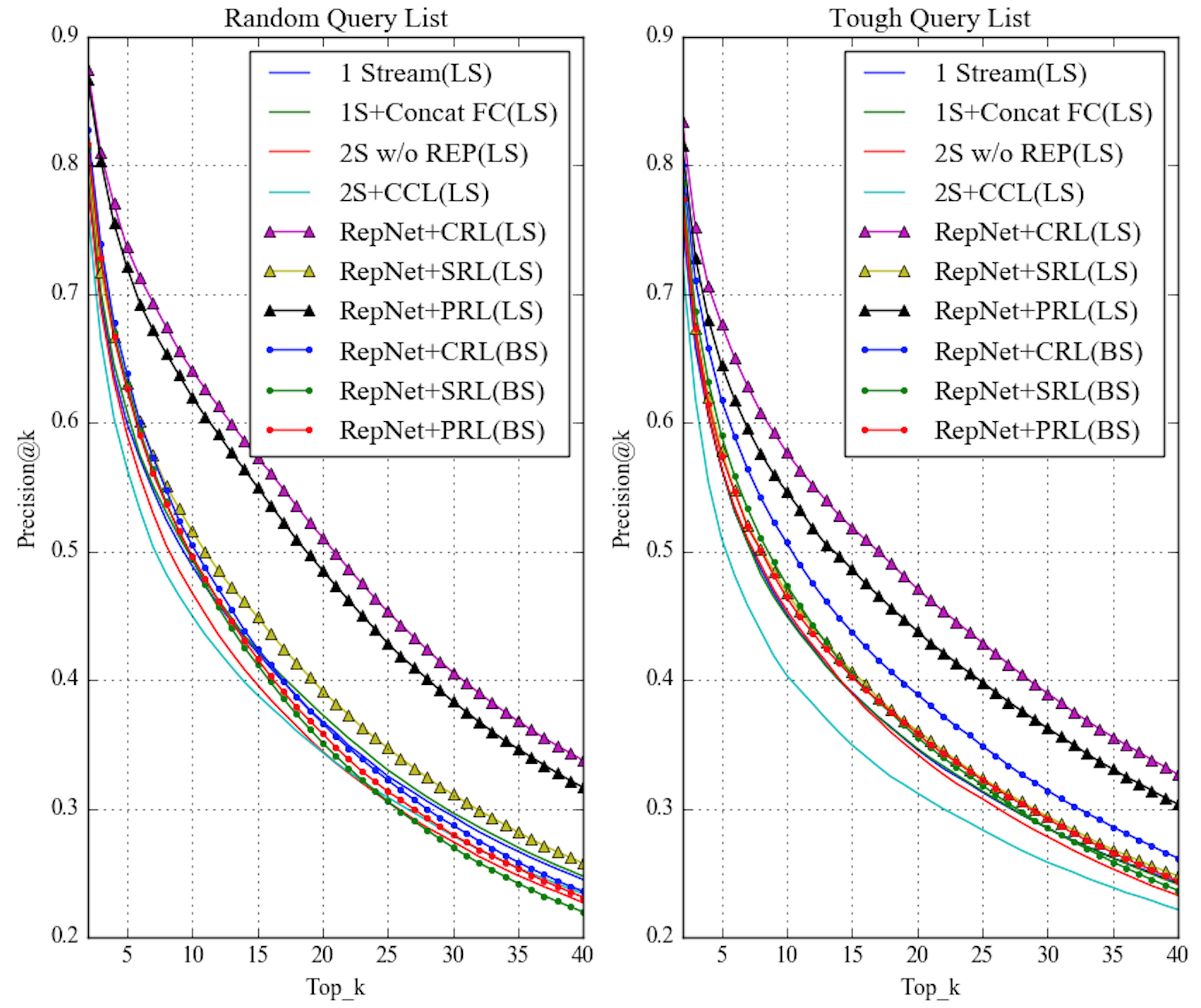}
  \caption{Precision curves for different models. LS and BS are abbreviations for linear search and bucket search.}
   \label{fig:pak}
\end{figure}

We compare our Repression Network with other multi-task learning frameworks mentioned at the beginning of this section. Since all models are constructed based on VGG and use constraints on both classification and similarity learning, we use only discriminative abbreviations to name them. 
MAP of all the retrieval tasks are listed in Table \ref{tab:map}. Two stream models outperform one stream models in bucket search which shows that splitting different constraints into different streams can help each of them to get better sub-feature. But with scale variations between sub-features, the performance of linear search, which needs to concatenates all the sub-features, for two stream models are greatly affected. 
RepNets, however, have remarkably better retrieval performance than all the other listed multi-task learning frameworks in linear search and CRL stands out from three repression layers for higher accuracy in both linear search and bucket search. It demonstrates that RepNet is effective for learning discriminative features for precise retrieval and Repression Layer can help balance the scale of sub-features. In addition, RepNet with PRL and CRL are the top two curves in Figure \ref{fig:pak}, which also strongly proves the effectiveness of our framework. 

Even though bucket search cannot have the same retrieval performance as linear search according to Table \ref{tab:map}, it can greatly reduce the retrieval time. In our experiments, we use features with dimension of 2348 (300 in $F_{SLS-3}$ and 2048 in $F_{ACS}$) for linear search, and features with dimension 300 for bucket search. The average retrieval time for linear search and bucket search are 0.213s and 0.009s respectively, which is a 24 times speed up for bucket search. Additionally, as shown in Figure \ref{fig:pak}, the precision@k curves of RepNet bucket search are no worse than those linear search curves of other models, which means the quality of the top part of the ranking list generated by RepNet bucket search is equal to those generated by linear search of other models. Moreover, the MAP of the RepNet bucket search with CRL on Random Query List is even higher than the linear search of other models.

\section{Conclusion}
In this paper, we propose a multi-task learning framework to effectively generate fine-grained feature representations for both attribute classification and similarity learning. A Repression layer is designed to keep the information learned from attributes classification being embedded into the feature representing specific details for each unique vehicle and also to balance the scale of weights in two FC layer streams. With a new attribute added to VehicleID dataset, extensive experiments conducted on that revised dataset prove the effectiveness of our framework as well as the repression layer --  higher image retrieval precision with linear search and less retrieval time with bucket search. These merits warrant further investigation of the RepNet on learning fine-grained feature representation such as introducing hash functions to generate binary features or splitting the convolutional groups into two groups. Besides, we can extend our framework into wider applications like face and person retrieval as well.


\ifCLASSOPTIONcompsoc
\else
\fi




\bibliographystyle{IEEEtran}
%

\bibliography{IEEEabrv,paper}

\begin{thebibliography}{10}
\providecommand{\url}[1]{#1}
\csname url@samestyle\endcsname
\providecommand{\newblock}{\relax}
\providecommand{\bibinfo}[2]{#2}
\providecommand{\BIBentrySTDinterwordspacing}{\spaceskip=0pt\relax}
\providecommand{\BIBentryALTinterwordstretchfactor}{4}
\providecommand{\BIBentryALTinterwordspacing}{\spaceskip=\fontdimen2\font plus
\BIBentryALTinterwordstretchfactor\fontdimen3\font minus
  \fontdimen4\font\relax}
\providecommand{\BIBforeignlanguage}[2]{{%
\expandafter\ifx\csname l@#1\endcsname\relax
\typeout{** WARNING: IEEEtran.bst: No hyphenation pattern has been}%
\typeout{** loaded for the language `#1'. Using the pattern for}%
\typeout{** the default language instead.}%
\else
\language=\csname l@#1\endcsname
\fi
#2}}
\providecommand{\BIBdecl}{\relax}
\BIBdecl

\bibitem{liu2016deep}
H.~Liu, Y.~Tian, Y.~Yang, L.~Pang, and T.~Huang, ``Deep relative distance
  learning: Tell the difference between similar vehicles,'' in
  \emph{Proceedings of the IEEE Conference on Computer Vision and Pattern
  Recognition}, 2016, pp. 2167--2175.

\bibitem{datta2008image}
R.~Datta, D.~Joshi, J.~Li, and J.~Z. Wang, ``Image retrieval: Ideas,
  influences, and trends of the new age,'' \emph{ACM Computing Surveys (Csur)},
  vol.~40, no.~2, p.~5, 2008.

\bibitem{krizhevsky2012imagenet}
A.~Krizhevsky, I.~Sutskever, and G.~E. Hinton, ``Imagenet classification with
  deep convolutional neural networks,'' in \emph{Advances in neural information
  processing systems}, 2012, pp. 1097--1105.

\bibitem{chai2013symbiotic}
Y.~Chai, V.~Lempitsky, and A.~Zisserman, ``Symbiotic segmentation and part
  localization for fine-grained categorization,'' in \emph{Proceedings of the
  IEEE International Conference on Computer Vision}, 2013, pp. 321--328.

\bibitem{zhang2013deformable}
N.~Zhang, R.~Farrell, F.~Iandola, and T.~Darrell, ``Deformable part descriptors
  for fine-grained recognition and attribute prediction,'' in \emph{Proceedings
  of the IEEE International Conference on Computer Vision}, 2013, pp. 729--736.

\bibitem{qian2015fine}
Q.~Qian, R.~Jin, S.~Zhu, and Y.~Lin, ``Fine-grained visual categorization via
  multi-stage metric learning,'' in \emph{Proceedings of the IEEE Conference on
  Computer Vision and Pattern Recognition}, 2015, pp. 3716--3724.

\bibitem{wang2014learning}
J.~Wang, Y.~Song, T.~Leung, C.~Rosenberg, J.~Wang, J.~Philbin, B.~Chen, and
  Y.~Wu, ``Learning fine-grained image similarity with deep ranking,'' in
  \emph{Proceedings of the IEEE Conference on Computer Vision and Pattern
  Recognition}, 2014, pp. 1386--1393.

\bibitem{chopra2005learning}
S.~Chopra, R.~Hadsell, and Y.~LeCun, ``Learning a similarity metric
  discriminatively, with application to face verification,'' in \emph{Computer
  Vision and Pattern Recognition, 2005. CVPR 2005. IEEE Computer Society
  Conference on}, vol.~1.\hskip 1em plus 0.5em minus 0.4em\relax IEEE, 2005,
  pp. 539--546.

\bibitem{chechik2010large}
G.~Chechik, V.~Sharma, U.~Shalit, and S.~Bengio, ``Large scale online learning
  of image similarity through ranking,'' \emph{Journal of Machine Learning
  Research}, vol.~11, no. Mar, pp. 1109--1135, 2010.

\bibitem{zhang2015embedding}
X.~Zhang, F.~Zhou, Y.~Lin, and S.~Zhang, ``Embedding label structures for
  fine-grained feature representation,'' in \emph{Proceedings of the IEEE
  Conference on Computer Vision and Pattern Recognition}, 2016, pp. 1114--1123.

\bibitem{schroff2015facenet}
F.~Schroff, D.~Kalenichenko, and J.~Philbin, ``Facenet: A unified embedding for
  face recognition and clustering,'' in \emph{Proceedings of the IEEE
  Conference on Computer Vision and Pattern Recognition}, 2015, pp. 815--823.

\bibitem{sun2014deep}
Y.~Sun, Y.~Chen, X.~Wang, and X.~Tang, ``Deep learning face representation by
  joint identification-verification,'' in \emph{Advances in neural information
  processing systems}, 2014, pp. 1988--1996.

\bibitem{yi2014learning}
D.~Yi, Z.~Lei, S.~Liao, and S.~Z. Li, ``Learning face representation from
  scratch,'' \emph{arXiv preprint arXiv:1411.7923}, 2014.

\bibitem{huang2015cross}
J.~Huang, R.~S. Feris, Q.~Chen, and S.~Yan, ``Cross-domain image retrieval with
  a dual attribute-aware ranking network,'' in \emph{Proceedings of the IEEE
  International Conference on Computer Vision}, 2015, pp. 1062--1070.

\bibitem{zhao2015deep}
F.~Zhao, Y.~Huang, L.~Wang, and T.~Tan, ``Deep semantic ranking based hashing
  for multi-label image retrieval,'' in \emph{Proceedings of the IEEE
  Conference on Computer Vision and Pattern Recognition}, 2015, pp. 1556--1564.

\bibitem{Simonyan14c}
K.~Simonyan and A.~Zisserman, ``Very deep convolutional networks for
  large-scale image recognition,'' \emph{CoRR}, vol. abs/1409.1556, 2014.

\bibitem{jia2014caffe}
Y.~Jia, E.~Shelhamer, J.~Donahue, S.~Karayev, J.~Long, R.~Girshick,
  S.~Guadarrama, and T.~Darrell, ``Caffe: Convolutional architecture for fast
  feature embedding,'' in \emph{Proceedings of the 22nd ACM international
  conference on Multimedia}.\hskip 1em plus 0.5em minus 0.4em\relax ACM, 2014,
  pp. 675--678.

\bibitem{knapp1978canonical}
T.~R. Knapp, ``Canonical correlation analysis: A general parametric
  significance-testing system.'' \emph{Psychological Bulletin}, vol.~85, no.~2,
  p. 410, 1978.

\bibitem{zeiler2014visualizing}
M.~D. Zeiler and R.~Fergus, ``Visualizing and understanding convolutional
  networks,'' in \emph{European conference on computer vision}.\hskip 1em plus
  0.5em minus 0.4em\relax Springer, 2014, pp. 818--833.

\end{thebibliography}




\end{document}